\documentclass[12pt]{article}
\usepackage{amsmath}
\usepackage{amssymb}
\usepackage{amsthm}
\usepackage{bm}
\usepackage{fancyvrb}
\usepackage{graphicx}
\usepackage{comment}
\usepackage{clrscode3e}
\usepackage{xurl}
\usepackage{tikz}	% For DIY graphics.
\usetikzlibrary{chains,shapes,positioning}

\usepackage[bottom=1in,top=1in,left=1in,right=1in]{geometry}

\newcounter{protocol}
\newcounter{msgcnt}
\renewcommand\themsgcnt{(M\arabic{msgcnt})}
\DeclareRobustCommand\stepmsgcnt{\refstepcounter{msgcnt}\themsgcnt}
\title{AI-Oracle Machines for Intelligent Computing\thanks{Copyright \copyright Jie Wang, 2024.}}
\author{Jie Wang\thanks{Richard Miner School of Computer and Information Sciences,
University of Massachusetts, Lowell, MA 01854, USA. Email: Jie\_Wang@uml.edu.} 
}
\date{}

\begin{document}	
\maketitle{}

% Add a bio for each author.  (These will appear at the end of the document.)
% Note:  Due to quirks in the way LaTeX wraps text around figures, only use \textbf or \textem for emphasis in the bios; no other formatting commands!  (Specifically, don't use \bf, \em, etc.)
%\AIMbio{sigai.jpg}{\textbf{The AI Matters Editors} may be contacted at \href{mailto:aimatters@sigai.acm.org}{aimatters@sigai.acm.org} with questions about the submission process for the AI Matters newsletter.  Complete submission instructions may be found on the AI Matters website.  If you note any bugs in the AI Matters template, please let the editors know.}

%\AIMbio{wang.jpg}{\textbf{Jie Wang} has been a Professor of Computer Science at the University of Massachusetts Lowell since 2001, where he served as department chair from 2007 to 2016. His research includes applied AI, data modeling, text mining, document engineering, algorithms and combinatorial optimization, medical computation, network security, and computational complexity theory. }

\begin{abstract}
We introduce the concept of AI-oracle machines for intelligent computing and outline several applications to demonstrate their potential. Following this, we advocate for the development of a comprehensive platform to streamline the implementation of AI-oracle machines.
\end{abstract}

\section{Introduction}

Oracle Turing Machines (OTMs) extend the computational limits of standard Turing machines by incorporating an oracle, enabling the exploration of relativized computability.
An oracle represents an external source of specific knowledge, capable of providing correct answers to any queries related to that knowledge. 

Replacing the oracle with a set of tangible AI models called \textsl{AI-oracle} results in a concrete OTM called \textsl{AI-oracle machine}. An AI model can be an in-house trained neural network, an out-of-the-box or fine-tuned pre-trained model such as a large language model (LLM), a large reasoning model (LRM), a large vision model (LVM), or any future advancements in AI technology. 

LLMs are capable of understanding and generating human language, acting as sources of contextual or domain-specific knowledge to assist in problem-solving. LRMs focus on logical reasoning, decision-making, and problem-solving abilities beyond text generation, making them valuable for complex computational tasks involving deep reasoning. LVMs tackle tasks involving image and video data, particularly those that require understanding and interpreting visual information.
At the time of writing, representative LLMs include GPT-4o, Claude 3.5, LLaMA 3, and Gemini 1.5; representative LRMs include GPT-o1; and representative LVMs include DALL-E 3 and Midjourney 6.

AI models generate answers to queries in a black-box manner, providing no guarantee of correctness or control over the outputs.
%even when equipped with self-correction capabilities, 
Prompt engineering is the primary mechanism for guiding AI models to produce the desired response, which may also include explanations of how the answer is derived, such as through the use of the chain-of-thought technique to demonstrate reasoning. These explanations, however, are not guaranteed to genuinely align with the underlying truth.

By combining conventional algorithmic techniques with AI models' vast knowledge bases, advanced text and visual understanding and generation capabilities, and inference power, AI-oracle machines can address these issues to a large extent. 

In a nutshell, AI-oracle machines achieve better control by decomposing complex tasks into manageable subtasks, guiding the formulation of queries through intermediate processing, and ensuring that each step aligns with predefined requirements. They use pre-query algorithms to form queries for completing subtasks toward achieving the original goal. Post-answer algorithms are then used to assess the accuracy of AI-generated responses by comparing them to user-provided reference materials, serving as the ground truth. These algorithms evaluate whether the entire response or only specific parts are valid, extracting relevant information to guide the next steps. This iterative process enhances the relevance, reliability, and accuracy of the final outcome.
%This approach of AI-oracle machines is particularly useful for applications where high accuracy and accountability are paramount, and where real-time solution generation is not a stringent requirement. 
%Examples include medical diagnosis, legal analysis, and complex decision-making systems.

AI-oracle machines extend the concept of LLM-oracle machines\footnote{J. Wang. LLM-oracle machines. 2024. \url{https://arxiv.org/abs/2406.12213v3.}} by expanding the oracle to include not only LLMs but also various other AI models, such as LVMs and LRMs, whether they are out-of-the-box pre-trained models, fine-tuned pre-trained models, or in-house trained models for specific tasks. This broader integration enables AI-oracle machines to handle more complex tasks, leveraging advanced capabilities in text processing, visual understanding, and logical inference, thus enhancing intelligent computing.

\section{An Overview of AI-Oracle Machines}

An AI-oracle machine $M$ is an OTM with a set of designated AI models as the oracle, denoted by $O_M$.
The input to $M$ is a tuple $(T,Q)$, where $T$ is a set of text or visual files representing the ground truth, and $Q$ is a task described in the form of interrogative or imperative sentences, with the possibility that $T$ may be empty.

In its computation, $M$ generates queries to complete specific tasks using the AI-oracle either adaptively, where subsequent queries depend on the results of previous ones, or non-adaptively, where queries are generated independently of earlier results. We refer to such a task as a \textsl{query-task}. 

Forming a query-task typically involves pre-query processing, which may include transforming data into a format suitable for querying or using algorithms to extract useful information from previous responses and derive intermediate results that aid in framing the query. It may also involve post-answer processing, which includes interpreting, aggregating, and validating the answer to ensure its correctness and relevance to the overall task.

Each query-task is in the form of a task description and a collection of attributes (which may be empty), including but not limited to the following:
\begin{enumerate}

\item Specific text or visual data extracted from $T$ to provide relevant context.

\item Specific contents extracted from one or more answers to previous subtasks.

\item Specific requirements to detail the desired outcome.

\item Specific examples to specify what the desired outputs should be like.

\item Specific constraints to outline what cannot be done or what conditions must be satisfied.

\item Specific method to validate the correctness of the answer to the subtask.
\end{enumerate}

For each query-task, $M$ forms a prompt based on the task description and attributes. It then selects an appropriate AI-model from $O_M$ to generate an answer. Depending on the validation result, $M$ may generate further query-tasks.
$M$ concludes the computation with the final answer to $Q$ once it reaches the halting state.

\section{Sample Applications}
\label{sec:applications}

We outline three applications using the frame\-work of AI-oracle machines. The first two have been implemented, while the third, which requires the guidance of neurosurgeons, represents a promising project in development.

In addition to the application examples we will outline, other applications include, but are not limited to, academic paper assessment, AI-driven sales agents, investment decision-making, financial report analysis, legal case analysis, and medical diagnosis. The possibilities are virtually unlimited.

\subsection{Summarizing a Specific Topic in an Article}

Suppose we want to summarize a specific topic within an article. While we can directly ask an LLM such as GPT-4o to generate a summary, we have little control over the content it generates. Instead, we can design an AI-oracle machine to achieve better control of the output content as follows:

\begin{enumerate}
\item Identify sentences in the article relevant to the topic.
\item Rank these sentences based on their importance and the diversity of sub-topics within the topic.
\item Select an optimal number of top-ranked sentences that align with the user's requirement for summary length. 

\item Prompt the underlying LLM to generate a summary based on the selected sentences within the context of the article.

\item Validate whether the generated summary, or parts of it, has adequately covered all significant aspects of the specific topic. It would also identify and remove any irrelevant or redundant content from the output, ensuring that only the most pertinent information is retained for the next step of the computation.
\end{enumerate}

%The summarizer provided by Doenba.ai is a concrete implementation of such an AI-oracle machine.

\subsection{Readability Assessment}

Asking GPT-4o directly about the pre-college grade level of a given article often results in an incorrect answer. Automatic readability assessment (ARA) of teaching materials for pre-college students at specific grade levels can be attempted by fine-tuning an GPT-4o. However, experiments conducted by a PhD student at UMass Lowell show that its accuracy falls short of the state-of-the-art (SOTA) method, which trains an SVC classifier using hundreds of linguistic features (LFs). Unfortunately, the SOTA method achieves only about 50\% accuracy for articles from the CommonLit Digital Library (CLDL) at \url{https://CommonLit.org}. 

The research constructs an AI-oracle machine to offer a new approach to improving ARA accuracy from 50\% to over 65\%, representing a 24\% improvement. %\cite{YuWang2024}. 
This method employs a local search algorithm supported by various fine-tuned LLM models, using a dataset of 1,654 articles collected from CLDL, which span grades 3 to 12 across 33 genres. 
%The fine-tuning process follows an 80-20 split for training and testing.

\textsl{The fine-tuning phase}:
\begin{enumerate}
\item Fine-tune GPT-4o to predict the genre of a given article, referred to as the \textsl{genre assessor}. 
\item Partition the dataset into subsets with similar genres to ensure that within each subset, there is a sufficient number of articles with a fairly even grade distribution.
Fine-tune GPT-4o separately for each subset to predict
a grade level for an article, referred to as the \textsl{grade assessor}.
\item Fine-tune GPT-4o to compare two articles and determine which one is more difficult, referred to as the \textsl{text comparator}
\end{enumerate} 

\textsl{The local search phase}: 

On an input article, the AI-oracle machine, with the oracle consisting of the
genre predictor, grade assessors, and text comparator,
 performs the following local-search algorithm:
\begin{enumerate}
\item Use the genre assessor to predict the genre of the input article. 
\item Use the corresponding grade assessor with respect to the genre partition that contains the predicted genre to compute a grade level of the input article, which serves as the starting point for the local search.
\item Use the text comparator to compare the input article with each article in a chosen reference set of articles in the training dataset for the predicted grade. Average the comparison scores to determine the search direction, whether to move up or down the grade levels. Continue adjusting the search based on these comparisons, and identify the final grade level when the local search concludes.
\end{enumerate}

\subsection{Neurosurgery Treatment Planning}

Brain tumors can be treated with gamma knives, a form of radiosurgery that delivers highly focused beams of gamma radiation, known as shots, to precisely target specific regions of the brain.
The coverage volume of each shot can be roughly visualized as a solid ball, available in several different sizes. 

An ideal treatment plan should use the fewest shots necessary to deliver just enough doses to cover the entire tumor, while minimizing exposure to healthy tissues, avoiding vital organs such as the eyes and sinus systems in the delivery pathway, and adhering to patient-specific treatment requirements and constraints. This presents a challenging task for relatively large tumors with irregular contours, as they require multiple shots to ensure the entire tumor region is adequately covered.

Computer-aided treatment planning systems aim to provide an initial solution to this optimization problem, assisting neurosurgeons in designing optimal treatment plans that meet clinical objectives while minimizing risks to healthy tissues and vital organs. This area has been the subject of extensive and intensive study, with significant research efforts focused on improving accuracy, efficiency, and patient outcomes.

Training a neural network model specifically for generating treatment plans is feasible, but such a model is unlikely to fully accommodate the unique conditions of each patient. Adjustments to the initial solution generated by such a model may therefore be necessary.
We can address this issue through an AI-oracle machine, for example, as follows:

\textsl{The fine-tuning phase}:
\begin{enumerate} 
\item Fine-tune an LVM to generate treatment plans by collecting 3D images of successful treatment plans for real patients, along with doctors' text descriptions---structured in a predefined format to ensure clarity and avoid ambiguity---detailing treatment requirements, patient conditions, and specific constraints.

\item Fine-tune an LLM to convert the neurosurgeon's natural language text descriptions of patient conditions, constraints, and treatment requirements into a predefined format for further processing.
\end{enumerate}

\textsl{The optimization phase}:
\begin{enumerate}
\item Upload the 3D images of the patient's brain and tumor, as well as the neurosurgeon's text specification outlining the treatment requirements and the patient's conditions and constraints.

\item Use the fine-tuned LLM to convert the text description into the required format.

\item Use the fine-tuned LVM to generate an initial treatment plan based on the uploaded 3D images and the formatted description.

\item Iteratively and interactively verify whether the current plan meets the treatment requirements and satisfies the constraints and conditions. If the plan is not satisfactory, adjust it by modifying parameters such as shot size, dosage, targeted location, delivery angles, or by adding or removing shots, guided by feedback from the neurosurgeon. Use the fine-tuned LVM to generate a more refined treatment plan in each iteration.
\end{enumerate}

\section{Developing an Implementation Platform}
The iterative steps of pre-query processing, query and answer generation, and post-answer processing are central to the design and implementation of an AI-oracle machine for a specific task. 

Each pre-query processing step uses custom algorithms to analyze both the current information and the previous answers generated by the AI-oracle, ensuring that the next query-task is appropriately framed based on the evolving context of the task. The current information may include user feedback and the selection of the most relevant data from user-provided reference materials. 
%This is followed by forming an appropriate prompt to obtain an answer from a selected AI model. 

Each post-answer processing step uses custom algorithms to assess the validity of the answer for the corresponding query-task or to determine which parts of the answer are valid. These valid portions are then used to advance to the next cycle of pre-query processing, query and answer generation, and post-answer processing. 
%Query-tasks may also be generated for completing the pre-processing or post-processing tasks.

While we can implement a specific AI-oracle machine as a standalone system, it is more desirable to develop a platform tool that facilitates the implementation of various AI-oracle machines 
across a wide range of complex applications. 

Creating such a platform would be more challenging than developing a RAG (retrieval-augmented generation) tool or a multi-agent RAG framework. The former relies on conventional retrieval algorithms to extract relevant information from reference materials based on user queries, while the latter facilitates collaboration between agents, each with distinct functions, working together to address complex tasks. The challenges arise from the need to integrate custom algorithms for pre-query and post-answer processing in the platform that supports iterative cycles of pre-query processing, query and answer generation, and post-answer processing.

%Each AI-oracle machine operates as an interactive algorithm involving AI models and user input, potentially throughout the entire computation process. Consequently, a platform tool for implementing AI-oracle machines can be considered a universal AI-oracle machine.

\section{Conclusion}

AI-oracle machines combine algorithmic rigor with the vast knowledge depositories and advanced inferencing capabilities of AI models. This fusion allows for more sophisticated intelligent computing, 
making them particularly suited for complex, high-stakes applications. The iterative, controlled nature of their computations enables these tasks, which demand precision and interpretability, to be addressed with transparency.

%\bibliography{AI-Oracle-Machine.bib}

%\bibliographystyle{apacite}

\end{document}